\documentclass[letterpaper]{article}
\usepackage{iccc}

\usepackage{times}
\usepackage{helvet}
\usepackage{courier}
\usepackage{graphicx}
\usepackage{url}
\usepackage{natbib}
\usepackage[final]{pdfpages}
\usepackage{graphbox,graphicx}

\author{Boyd Branch\textsuperscript{*} \enspace Piotr Mirowski\textsuperscript{*} \enspace Kory Mathewson \\ Improbotics
\And \qquad \qquad \qquad \qquad Sophia Ppali \enspace Alexandra Covaci \\ \qquad \qquad \qquad \qquad \qquad University of Kent 
}

\begin{document} 

\title{Designing and Evaluating Dialogue LLMs for Co-Creative Improvised Theatre}

\maketitle
\begin{abstract}
Social robotics researchers are increasingly interested in multi-party trained conversational agents. With a growing demand for real-world evaluations, our study presents Large Language Models (LLMs) deployed in a month-long live show at the Edinburgh Festival Fringe. This case study investigates human improvisers co-creating with conversational agents in a professional theatre setting. We explore the technical capabilities and constraints of on-the-spot multi-party dialogue, providing comprehensive insights from both audience and performer experiences with AI on stage. Our human-in-the-loop methodology underlines the challenges of these LLMs in generating context-relevant responses, stressing the user interface's crucial role. Audience feedback indicates an evolving interest for AI-driven live entertainment, direct human-AI interaction, and a diverse range of expectations about AI's conversational competence and utility as a creativity support tool. Human performers express immense enthusiasm, varied satisfaction, and the evolving public opinion highlights mixed emotions about AI's role in arts.

\end{abstract}

\section{Introduction}
\begin{figure}
  \includegraphics[width=\columnwidth]{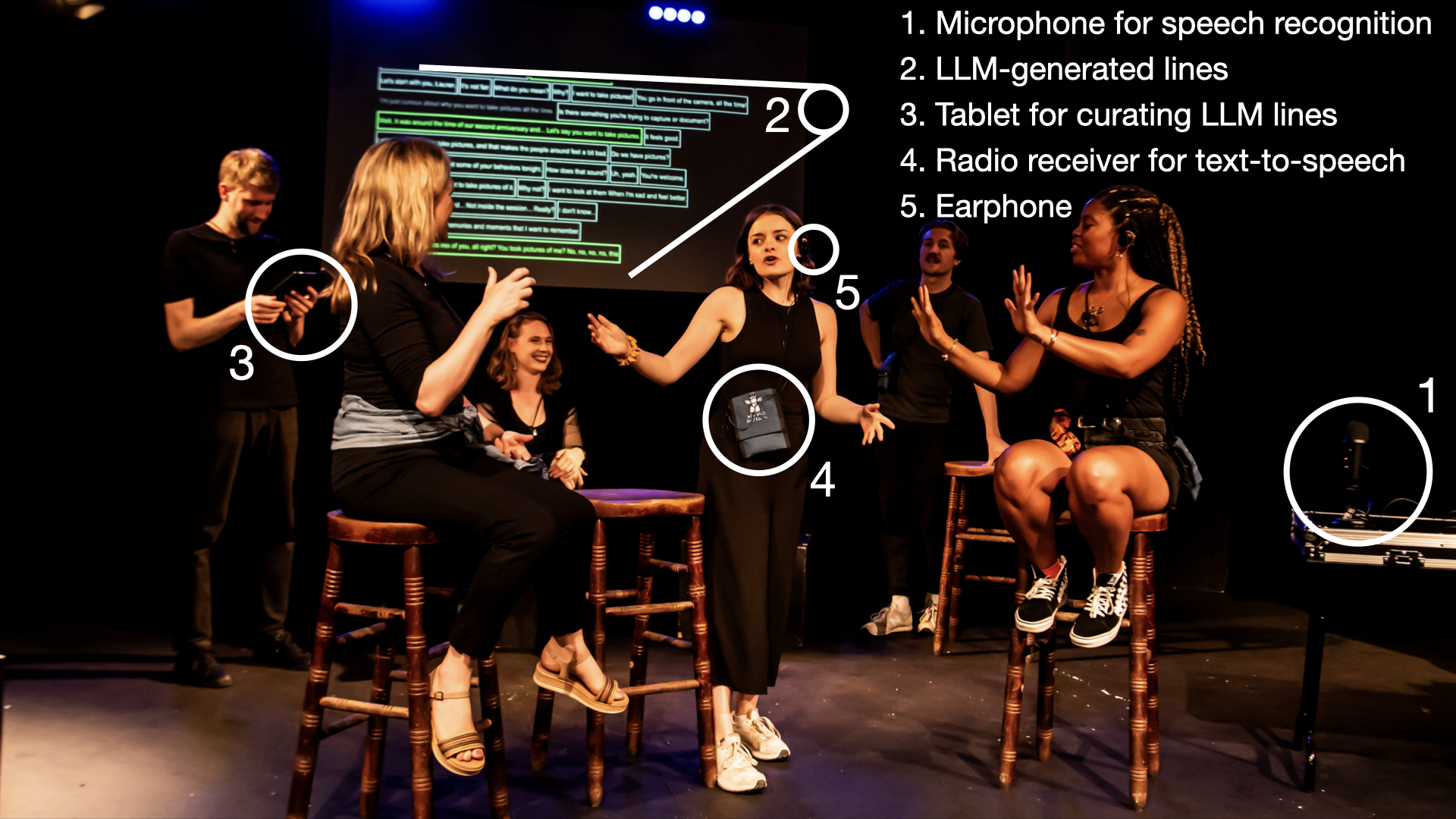}
  \caption{Cast of Improbotics performing AI-based improv theatre. The \emph{``Cyborg''} is wearing an earphone connected to a radio system that receives text-to-speech from LLM-generated lines (and curated by an \emph{operator}). The LLM is prompted by speech recognition. Photo: Lidia Crisafulli.}
  \label{fig:humanloop}
\end{figure}

This case study examines the process of designing and staging chatbots to perform improvisational theatre alongside a cast of human actors (Improbotics\footnote{\scriptsize\url{improbotics.org}}), conversing naturally in \emph{Multi-Party Chat} in front of live audiences.
We choose improvised theatre \citep{johnstone2014impro} as a challenging testbed for human-machine co-creation enabled by the ability of language models to respond ``intelligently'' to performing arts scenarios involving both dialogue with other performers and interaction with audiences.
The show builds upon prior work in improvised theatre and comedy with physical robots \citep{bruce2000robot}, chatbots \citep{mathewson2017improvised,mathewson2017turing,cho2020grounding}, story generators \citep{branch2021collaborative}, comedy roast generators by company \emph{ComedyBytes}\footnote{{\scriptsize \url{comedybytes.io}}}, live machine translation \citep{mirowski2020rosetta}, joke generation \citep{toplyn2022witscript,goes2023gpt} and employs the conceit of a human actor (the \emph{``Cyborg''}) taking their lines from a chatbot, similarly to \citep{mathewson2018improbotics,loesel2020digital} and to the show \emph{Yes, Android}\footnote{{\scriptsize \url{blogto.com/events/yes-android-comedy-robots-toronto}}}. Note that none of these works explored the specific solution to multiparty dialogue interaction where the AI simultaneously interacts with multiple improvisational actors on stage.

We deployed a conversational agent powered by three different Large Language Models (LLMs)\footnote{Chat GPT 3.5 \citep{openai2023gpt} (OpenAI), PaLM 2 \citep{anil2023palm} (Google), Llama 2 \citep{touvron2023llama} (Meta)} to improvise with a company of professional actors for 26 different audiences during the 2023 Edinburgh Festival Fringe\footnote{\scriptsize \url{https://www.edfringe.com/}}. During the run of the show, we conducted surveys with both cast and audience members to examine general perceptions of conversational AI, and to understand the in-situ impact of AI on creativity, performance expectations when staging AI, and anxieties around robots. With our work we further demonstrate a participatory design model for developing AI tools for the performing arts
and examine how design choices impact its abilities to meaningfully contribute to group conversations. 

Our study was formed in response to the recent advancements in natural language computing and conversational AI which have highlighted the potential of ubiquitous AI participating in human social and creative lives \citep{youssef2022survey,lim2021social}. While current conversational AI applications mostly focus on single-user text-based dialogue, researchers are interested in the next frontier of Multi-Party Chat (MPC) AI \citep{traum2003issues,kirchhoff2003directions,poria2018meld,zhu2022multi,wei2023multi} that can intelligently respond not only to dialogue, but also to the physical and social context of the conversation. Issues in multi-party dialogue include speaker and addressee identification, turn and conversation thread management or establishing a common ground among multiple participants \citep{traum2003issues}. MPC is already prominent in online social gaming research and development where non-player characters (NPCs) react to multiple online players \citep{urbanek-etal-2019-learning} but it is also relevant for speculative AI
applications to human social, cultural, and political life. 

Various technical, psychological and social factors affect not only the feasibility of deploying robots in social contexts, but also their ``acceptability by humans as partners in the interaction \citep{youssef2022survey}.'' 
With some exceptions, including staging robots in theatre performances \citep{chikaraishi2017creation,nishiguchi2017theatrical}, Human Robot Interaction (HRI) studies also generally take place in laboratory settings.
Our case study was designed to introduce an application of MPC AI in a theatre space, highlighting real-world challenges of both social robotics and of conversational AI, and capturing human reactions to the system in action. Using \emph{the theatre as a laboratory to study how actors and audiences respond to the presence of a real-time performing AI}, our work answers ``a call for a more integrative approach when investigating HRIs'' that specifically takes into account ``the cultural context of the experiment'' \citep{lim2021social}, and puts participants in direct contact with the physically present robots.

Until recently, public perception of robots has been shaped by media more than by direct experiences with AI \citep{haring2014perception}, fostering both over- and under-estimation of AI's ability and utility by the public. Our findings demonstrate how \emph{both interacting with, and observing an AI being deployed in a social context, influences the perception of AI capability} as well as the motivation for interacting with it, and the ability to relate to it. 
Our case study is also an examination of participatory and user-centered design demonstrating how \emph{user-in-the-loop design compliments human-in-the-loop AI research}. We find that such participatory models of inquiry can tackle the intertwined technical and psychological challenges around the subjects of creativity, human computing, and robotics interaction.

The following sections outline our method for deploying and staging MPC LLMs around the unique limitations and opportunities of live theatre, our development of improvisational games designed to challenge and explore the limits of state of the art conversational agents, and the results of our audience and performer surveys that highlight how our design choices impacted perceptions of AI.

\section{Staging AI in Theatre Festival Performances}
Multi-party dialogue live on stage with a chatbot presented numerous challenges for the developers, researchers, and artists involved.
While in principle modern LLMs can track multiple conversational agents, they need a speech recognition system with multiple microphones to identify speakers, making it impractical for ``Fringe'' theatre productions with short theatre get-in and get-out times. Second, capturing speech does not account for the physical aspects of communication on stage, from gestures to tone of voice; missing the full context of the conversation. Third, design decisions must be made about the timing of that chatbot's responses.

\begin{figure}
 \centering
 \includegraphics[width=\columnwidth]{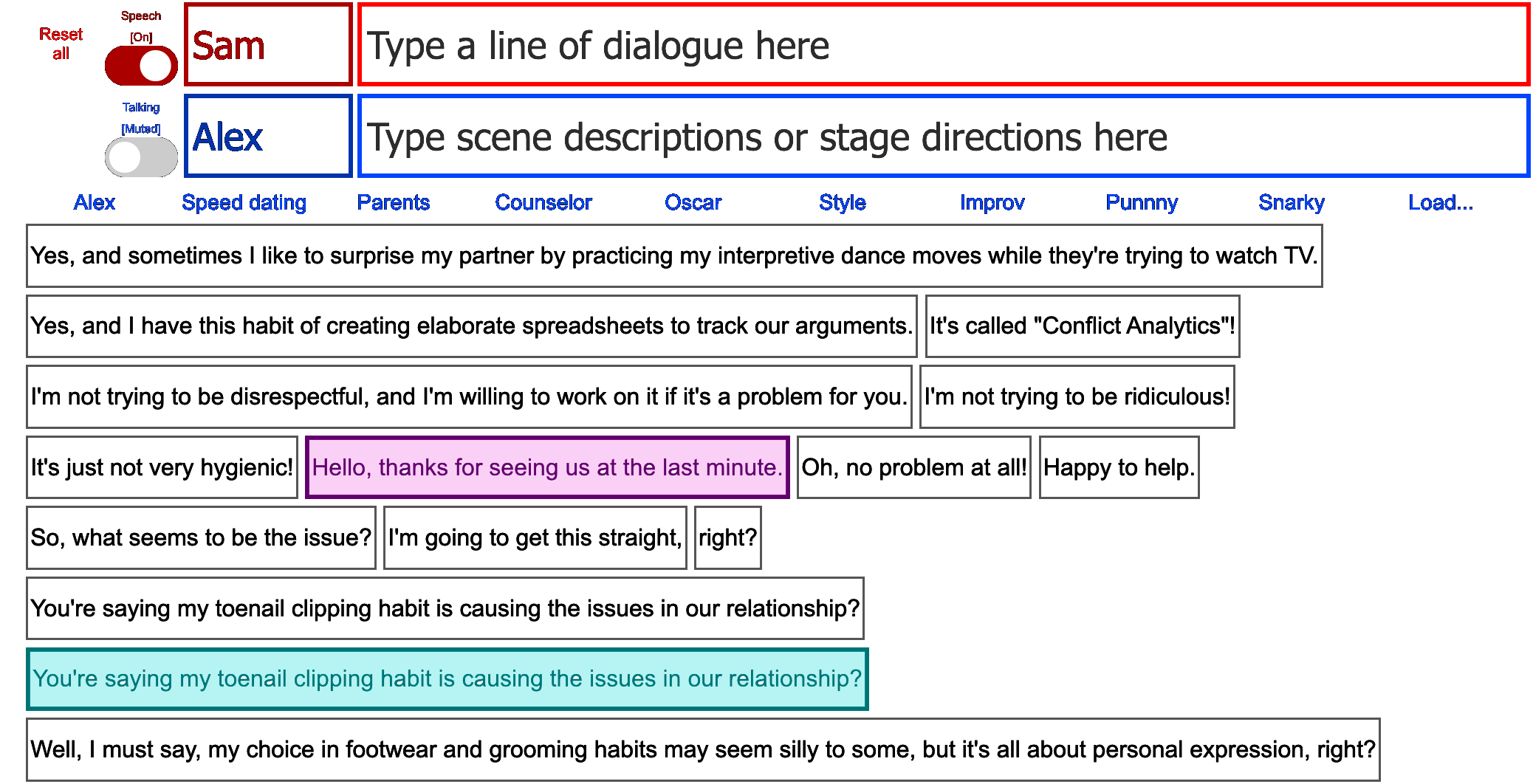}
 \caption{Screen capture of the AI \emph{Operator} interface. At the top (in red) is the input box for human character's name and for lines of dialogue (this input box serves as a backup in case speech recognition does not work properly). Below (in blue) is the input box for the AI character's name and for scene context metadata, typed by the operatore. Below are several buttons to rapidly input scene-specific prompts such as ``getting therapy'' or ``behaving in a sarcastic way''. The interface then shows multiple lines: AI-generated lines are in black, speech recognition lines are in pink, and the curator-selected lines are in cyan.}
 \label{fig:interface-operator}
\end{figure}

\begin{figure}
 \centering
 \includegraphics[width=0.8\columnwidth]{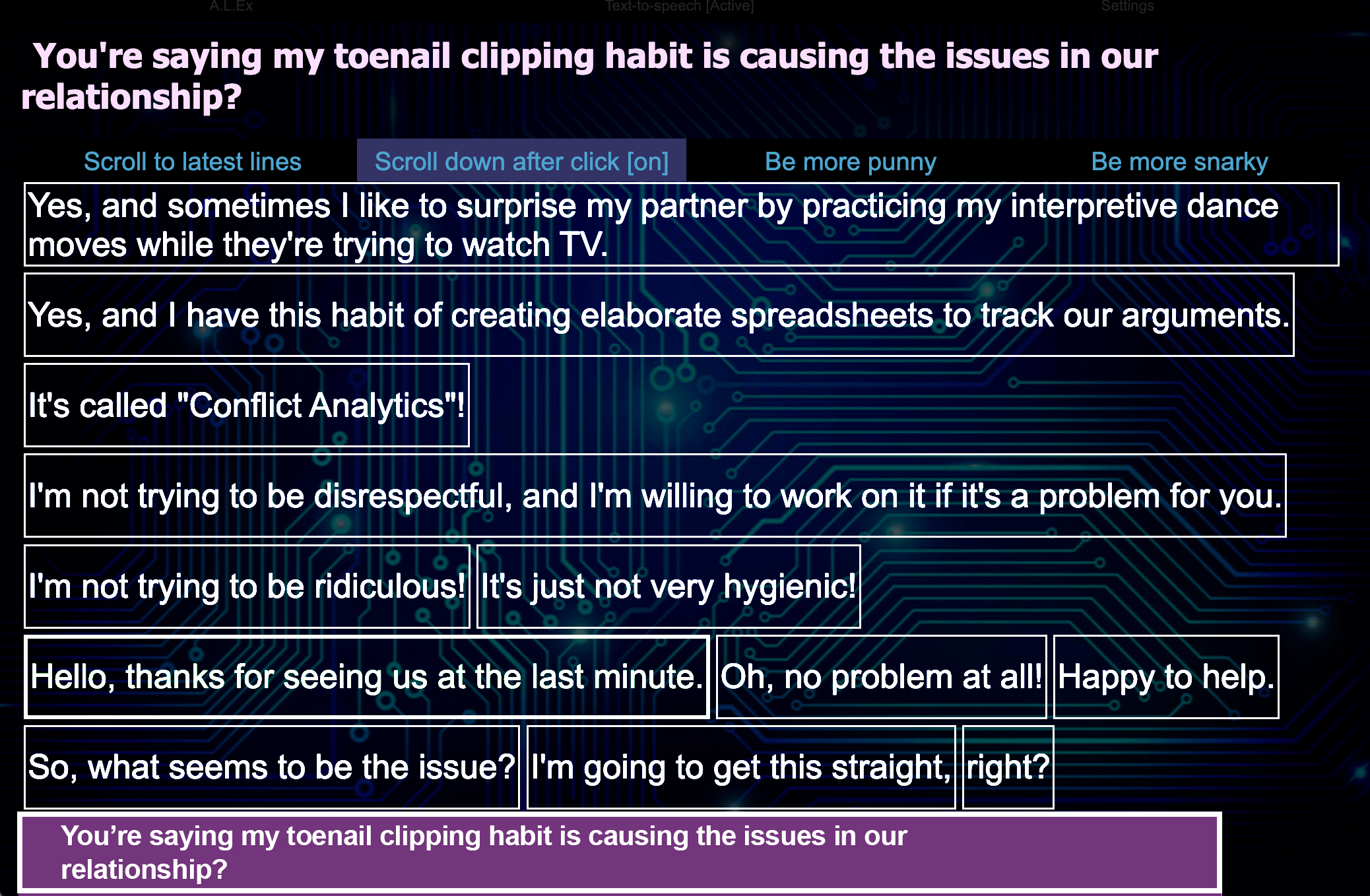}
 \caption{Screen capture of the AI \emph{Curator} interface, on a tablet. The latest speech recognition result is visible on top. Immediately below are buttons to scroll down to the latest AI-generated line, or to quickly input metadata (``more punny'' or ``more snarky'') for the language agent. Generated lines are in white and curator-selected lines in violet.}
 \label{fig:interface-curator}
\end{figure}

\begin{figure*}
    \centering
    \includegraphics[width=.75\textwidth]{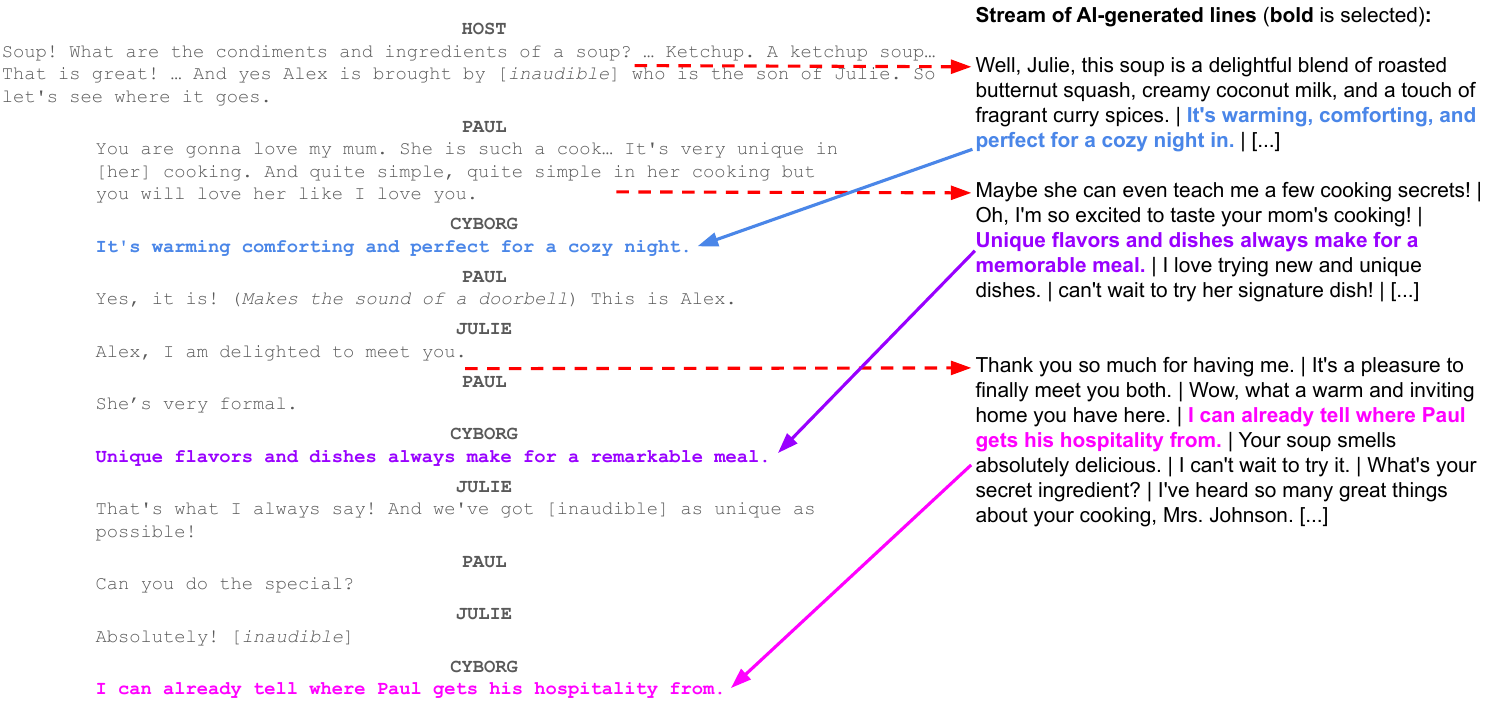}
    \caption{Dialogue from an improvised scene between Paul, Julie (his mother) and the Cyborg (Paul's date) where the Cyborg meet's Paul's parents, with suggestion: \emph{ketchup soup}. In this dialogue extract (recorded using speech recognition), the Cyborg says 3 lines, marked with solid blue, violet and pink arrows: ``It's warming, comforting, and perfect for a cozy night in'', ``Unique flavors and dishes always make for a memorable meal'' and ``I can already tell where Paul gets his hospitality from''. All LLM-generated lines are shown on the right side of the figure as a sequence of lines we call the \emph{AI Stream}. Red dashed arrows show which dialogue sentence triggered which LLM-generated line.}
    \label{fig:dialogue}
\end{figure*}

To address these challenges, we developed a system that combines continuous speech recognition (see Appendix for implementation details) and a \emph{human-in-the-loop curation system} to allows an \emph{Operator} to type contextual metadata for the LLMs, supplementing context from the live speech\footnote{Scene dialogue was recorded using a microphone and speech recognition system, resulting in numerous transcription errors.} captured on stage. We also engineered
prompts to influence the LLMs' style of response (see Fig. \ref{fig:interface-operator}). A second performer (\emph{Curator}) selected in real time the best response from a ``stream of consciousness''-like set of responses continuously generated by the LLMs (see Fig. \ref{fig:interface-curator}). Figure \ref{fig:dialogue} displays the AI's contributions in a short improv scene dialogue, with only a small subset selected by the curator and sent via text-to-speech and earpiece to the \emph{Cyborg} performer.

Co-creative storytelling abilities of AI have been initially explored using a custom LLM based on recurrent neural networks \citep{mathewson2017turing,mirowski2019human} trained on movie and television dialogue \citep{lison2016opensubtitles2016}. Their early model struggled with longer narrative threads and handling multiple speakers. OpenAI's release of GPT-2 \citep{radford2019language} with a Transformer architecture in 2019, introduced a much more powerful LLM that allowed for performance of longer narrative scenes \citep{mirowski2020rosetta,branch2021collaborative}. Nevertheless, multi-party dialogue performance remained underwhelming. In an improvement over their work, and experimenting with GPT-3 \citep{brown2020language} and subsequent models \citep{openai2023gpt,anil2023palm,touvron2023llama}, we found that given sufficient prompting and context, GPT-3 and later models appeared capable of producing nuanced, varied responses that addressed more than one participant in the scene. This led us to explore new games and formats for improvisational theatre that involved more than one scene partner.

\begin{figure}
 \centering
 \includegraphics[width=0.8\columnwidth]{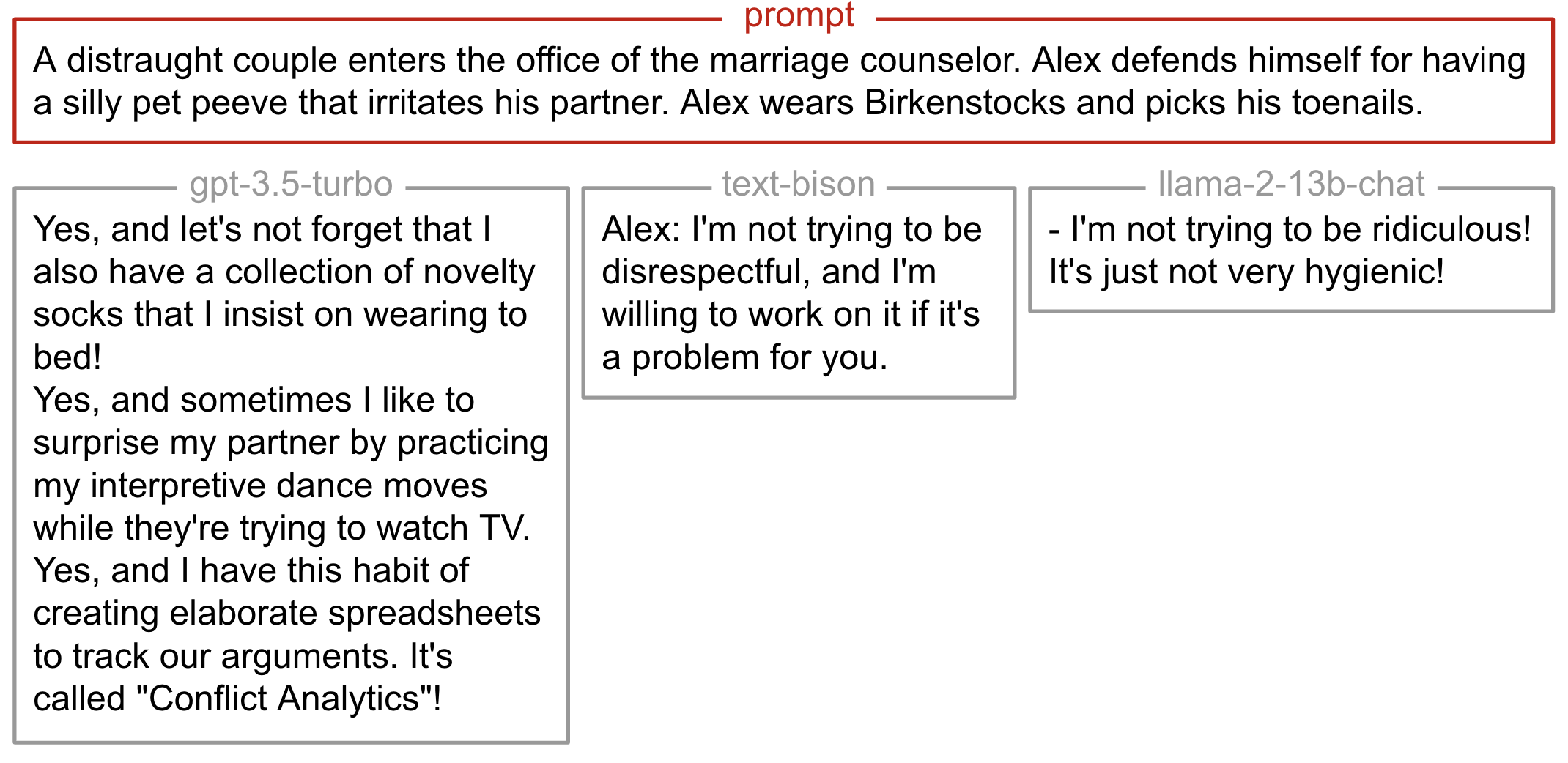}
 \caption{Example of instruction prompt and results generated for three different LLMs.}
 \label{fig:prompts-results}
\end{figure}

For the 2023 Edinburgh Festival Fringe, we developed a system that prompted multiple LLMs simultaneously while incorporating live stage dialogue. We adopted a participatory iterative design approach, with actors trying the system and providing regular feedback about improvements to support role-playing with AI. Subsequently, through several cycles of experimentation and feedback, we developed a series of improv games that appeared to allow improvisers to perform comfortably while also allowing us to continue testing the capacity of LLMs to handle a variety of complex multi-party scenes. The success and consistency of our system, and the positive performer feedback gave us confidence to present the show as part of the Fringe Festival for 26 consecutive performances to new audiences each night.
 
Over the course of the festival, we performed the show for over 1750 people across 26 unique performances with various groups of 20 different improvisers. Using audience and performer surveys, logs of LLM interactions, as well as close observations of the performances, we examine various practices of prompt engineering, UI/UX design around human-in-the-loop AI curation, audience experience and identification with the presence of AI on stage as part of an ensemble, as well as real time multi-party co-creation of narratives with AI and human improvisers.

The framing of our show was detailed in the festival program, and on-stage at the beginning of the performance, making clear to the audience that AI assisted the creativity of the improvisers \citep{colton2011computational}.

\section{Participatory \textit{Human-in-the-Loop} Design}
Performing at the Edinburgh Festival Fringe presented a unique opportunity to test new features of state-of-the-art LLMs for audiences who, for the first time, were generally knowledgeable about how LLMs functioned \citep{kuzior2022cognitive} and to collect data on general perception of AI. Moreover, most of our performances had been for short runs of 2-3 consecutive shows: performing numerous shows back-to-back allowed us to engage in concentrated participatory design model of technology, putting audiences and actors directly \textit{in the loop}, together with our development team, working to make new technology entertaining.

\subsection{Iterative Design with the Theatre Cast}

Staging AI in a real-time theatrical setting presents technical challenges. Of primary importance for the stage is the question of its embodiment.
Early Improbotics's acts involved both a small 
physical robot\footnote{EZ-Robot {\scriptsize\url{ez-robot.com}}, Aldebaran {\scriptsize\url{aldebaran.com/en/nao}}} that moves as it ``speaks'', as well as a human performer receiving lines from the chatbot via an earpiece and delivering them with physical and emotional interpretation.
We refer to the latter as ``\emph{Cyborg}''.

Previous work had largely focused on dialogue-based games that operated as a kind of \textit{``Turing Test''}~\citep{mathewson2017turing} to see how well a chatbot could perform a human activity, and on challenging situations for a robot competing with humans.
That concept kept the AI as an object of the scene, specifically the object at the end of the implied joke that success would only be accidental or as a result of skilled improvisers who could weave order back into the chaos it presented. 

Prior to adopting ChatGPT-3.5 \citep{openai2023gpt}, PaLM 2 \citep{anil2023palm} or Llama 2 \citep{touvron2023llama}, chatbots on stage usually presented non sequiturs and absurd sounding responses to the human dialogue, and much of the humour came from watching the improvisers try to make sense of what was being said \citep{loesel2020digital}.
With the latter models, we observed that the \emph{Cyborg} performer could provide reasonable and appropriate sounding responses which seemed to also deprive the scenes of humour. Subsequently our company members began debating about a possible inverse relationship between language capabilities of LLMs and what makes them funny on stage. Some argued that delight and humour derive from the absurdity of responses by the AI, and that better language models would lessen the humour on stage. Others argued that a different kind of interest and delight will emerge with increasingly intelligent robots that no longer are the object of the joke. The lively discussion of the topic inspired us to develop new formats and games that would allow the AI to function within scenes, rather than as the object of the scenes for actors. The following sections details the iterative development of the game format, and how games were developed to support co-creativity with AI scene partners more than presenting just another ``Turing''-style test for the audience.

\subsection{Designing for the Theatre Audience}

Besides testing the ability of our modern LLM models to handle MPC, we also needed to take into account the entertaining aim for the show: audiences were not buying tickets to provide feedback about the ability of LLMs to handle MPC dialogue. This meant we needed to find a show format where a successful performance would not be tied solely to the AI performance, but to the ensemble cast performance that included the AI as an equal cast member.

We therefore designed new games we hoped would allow us to assess the collaborative skill of AI and its added value to a live theatre experience.
In the following, we slightly anthropomorphize the AI
and we distinguish between the \emph{Robot} when the AI controls a robot, and \emph{Cyborg} when the AI provides lines for a human actor. In the following games, the main feedback mechanism and success criterion corresponded to the perceived amount of laughter, as used in similar studies with robot comedians \citep{vilk2020comedians}.

\subsubsection{Speed Dating} To start the show and introduce the audience to the skills of the AI, we developed an opening game of ``Speed Dating'', where the Robot was tasked with going on multiple dates with different characters in an attempt to find a mate. A human improviser in the same situation is capable of quickly adapting their behavior to the offers made by the various dates, in turn revealing various aspects of their own character over time, which eventually broadcast the unique likes and dislikes that one of the ensemble of dating characters will match with. Putting the Robot in the main role would allow us test how well variations of likes and dislikes might emerge from interacting with multiple contrasting personalities. We observed that the AI could not convincingly provide consistent dating criteria. Instead, the success of the game for the audience simply depended on how the Robot responded to the outlandish offers being made by the human improvisers.

\subsubsection{Wedding Speech} While ``Speed Dating'' looked at how well the AI could rapidly respond to multiple speakers, one line at a time, ``Wedding Speech'' was designed to explore the LLMs' ability to incorporate multiple inputs from both audience members and cast members into a longer narrative. For this game the Cyborg is required to give an impromptu speech at the wedding of a former lover, speech prompted both from the dialogue of a preceding scene, as well as suggestions from the audience about secrets that may get revealed. The game therefore tested how well the AI would be able to take disparate threads of a potential story and weave them together into a coherent as well as entertaining speech that would ``surprise'' the audience. 

\subsubsection{Couples' Therapy and Meet the Parents} Two other games involving the Cyborg going to ``couples' therapy'' and ``meet the parents'' of a partner for the first time, provided scenes where the Cyborg would need to converse with two different people who had different needs, expectations, and desires related to the Cyborg. With these games we hoped to provide both audiences and actors with a wide range of types of social encounters with the Cyborg, which allowed for its successes and failures in a given scene to be evenly considered across a gamut of scenes where it would sometimes be the focus of attention, or sometimes just a supporting role.

\subsubsection{Hero's Journey} We developed ``Hero's Journey'' \citep{campbell2008hero} as the penultimate challenge to explore the range of long-term memory and ability to distinguish itself amongst a group of improvisers constantly changing roles. With this game, the Cyborg is given an undesirable job (point of departure), and an aspirational job (destination); working with an ensemble of 4-5 human performers, the Cyborg must overcome obstacles on their journey to the career of their dreams. This is the most technically challenging role for the Cyborg since it encounters repeating scene partners, each presenting as obstacles or allies in achieving a change in job and status. Like our other games, the format does not depend exclusively on the Cyborg driving the scenes, but allows it to be a relatively passive protagonist encountering a slew of characters guiding it to the destination.

\subsubsection{Improvised TED Talk and Movie Pitch} Our show used two additional game formats, namely ``Improvised TED Talk'' (relying on AI-generated PowerPoint slides, unbeknownst to the actor) \citep{winters2019automatically} and ``Movie Pitch'' (where image generators were used to output, in real time, images illustrating or disrupting an improvised film pitch). These games leveraged AI for idea generation.

\subsection{Designing AI Curation Interfaces}

With this show format we hoped to be able to simultaneously test the abilities of new LLMs, and to hedge our bets on being able to present an entertaining show. While we felt confident in the development of improv games, appropriately prompting LLMs on stage raised another challenge. Speech-to-text does not provide the LLMs with sufficient relevant context, and without specific prompt engineering, the LLMs tend toward ``honest, helpful and harmless'' information giving \citep{askell2021general,bai2022training} rather than chit-chat helpful for role-playing \citep{hendry2023you}. Previous studies of MPC NLPs focused their studies on specific scenarios with predetermined roles 
\citep{Zhang2023}.

In improvised theatre \citep{johnstone2014impro}, every scene is made up on the spot, meaning that predefined roles cannot be assigned. We thus developed a system that allows an operator to type in metadata to a given scene to ``steer'' the AI agent towards a personality or agenda within the scene. The speed at which scenes take place however, makes such live-inputting a daunting tasks. Moreover, we could not give these instructions via voice input, since voice was already used for scene dialogue.

Eventually, we built buttons into the user interface to allow the operator to direct the AI agent to provide more provocative and emotionally charged responses. After some trial and error in rehearsal we found that prompting the LLM to respond in the role of a ``sarcastic'' and ``pithy'' friend in a addition to the scene specific prompts injected a bit of playful conflict that could humanise the responses if used on occasion. The LLMs were also remarkably adept at making puns, which was useful for the scenes \textit{when used in moderation}. Furthermore, we tested giving more scene-specific prompts to help the LLM perform with stylistically different responses. For example, instructions such as ``try to repair your relationship by pointing out your partner's flaws to the therapist'', resulted in better scene-specific ``playful conflict'' that could be resolved over time. Subsequently, we created buttons in our UI for the LLMs to be ``more snarky'' or ``more punny'', and inject relationship contexts with buttons for ``reminiscing with loved ones'', or ``getting therapy.'' Such prompts would be interjected along with the live real-time audio captured from an onstage microphone and processed with a speech-to-text model (see Appendix) \citep{radford2023robust}, which became a stream of prompts to LLMs.

Finally a curator was given a tablet to select from the stream of lines constantly generated by the LLMs (see Fig. \ref{fig:humanloop} for the illustration of the setup). The person selecting the lines would not know which LLM had served that response, allowing us to later analyse if there were any preferences for specific models.

\subsection{Audience and Actor Surveys After Performances}

To capture the impact of our iterative design choices (such as updates to the line curation UI) as well as to understand how actors and audiences experienced the show, we surveyed both groups after each performance,
with approval from the ethics governing board at the University of Kent\footnote{\scriptsize \url{kent.ac.uk/research-innovation-services/research-ethics-and-governance}}. 
Survey answers from both audiences and actors were collected anonymously via an online Qualtrics questionnaire; the links were shared via email (actors) or QR code shown before and after the show\footnote{The survey invitation was: ``Help us do real science! After the show, answer a 5-minute research survey on human-computer interaction.
Research project run by Dr. Boyd Branch and Dr. Piotr Mirowski. Ethics approval: University of Kent.'' 
Consent and privacy notice are detailed in the Appendix.} (audiences). To proceed with the online survey, participants needed to read a short statement about the study and to consent to the use of their data. 

\subsubsection{Survey Design}
Our performances took place in the context of a large theatre festival with dozens of different shows taking place each hour, and a 30-minute turn-around period between shows. Accordingly, we anticipated difficulty collecting a high volume of responses per show for both audiences as well as actors who needed to exit quickly. We chose to exclude collecting demographics data to shorten the surveys, and due to the likely high margin for error for collecting the actual demographics of the audience. Survey questions were exclusively focused around the primary questions of the experience of watching an AI perform on stage.

\subsubsection{Data}
Audience surveys accumulated 150 unique individual responses, and actors surveys 21 individual responses over the course of 26 shows. On average, 67 audience members attended each show (5.8 survey responses per show). The quick turn-around of the show (5 minutes fo audience get-in and get-out) impacted survey responses as only the first 5 questions were answered by all (150), and the subsequent 15 questions were answered on average by 109 unique individuals. As stated above, we did not collect demographic data in either actor or audience surveys. We noticed audiences of mixed genders, ages (including young teenagers), ethnicity and nationality, and overall representative of typical Edinburgh Festival Fringe attendees.

Audience questions were designed to capture experiences by participants related to how much they anthropomorphised the AI, their general attitudes about AI before the show, and any change in sentiment after the show. Multiple choice and open-ended questions were adapted from earlier studies of anthropomorphism, attachment, and scales for AI authenticity and AI social interaction conducted around the use of companion AI \citep{pentina2023exploring}. Actor surveys were focused on examining the experience of both performing with and as a Cyborg (adapted from previous work \citep{mathewson2018improbotics}) and open-ended questions designed to capture a holistic understanding of the performer experience with AI. The complete list of survey questions can be found in the Appendix. 

The following results are presented for discussion rather than to make statistically significant claims about audience perceptions of robots on stage. Collecting data ``in the wild'' of a theatre festival presented too many variables for us to make objective claims, besides our anticipation of the low volume of responses per show. The survey data therefore is used to inform qualitative evaluation of both the technical and performative aspects of the show, and provide insights for future research on robots in live performance.

\section{Post-Performance Analysis}
\begin{figure}
 \centering
 \includegraphics[width=0.49\columnwidth]{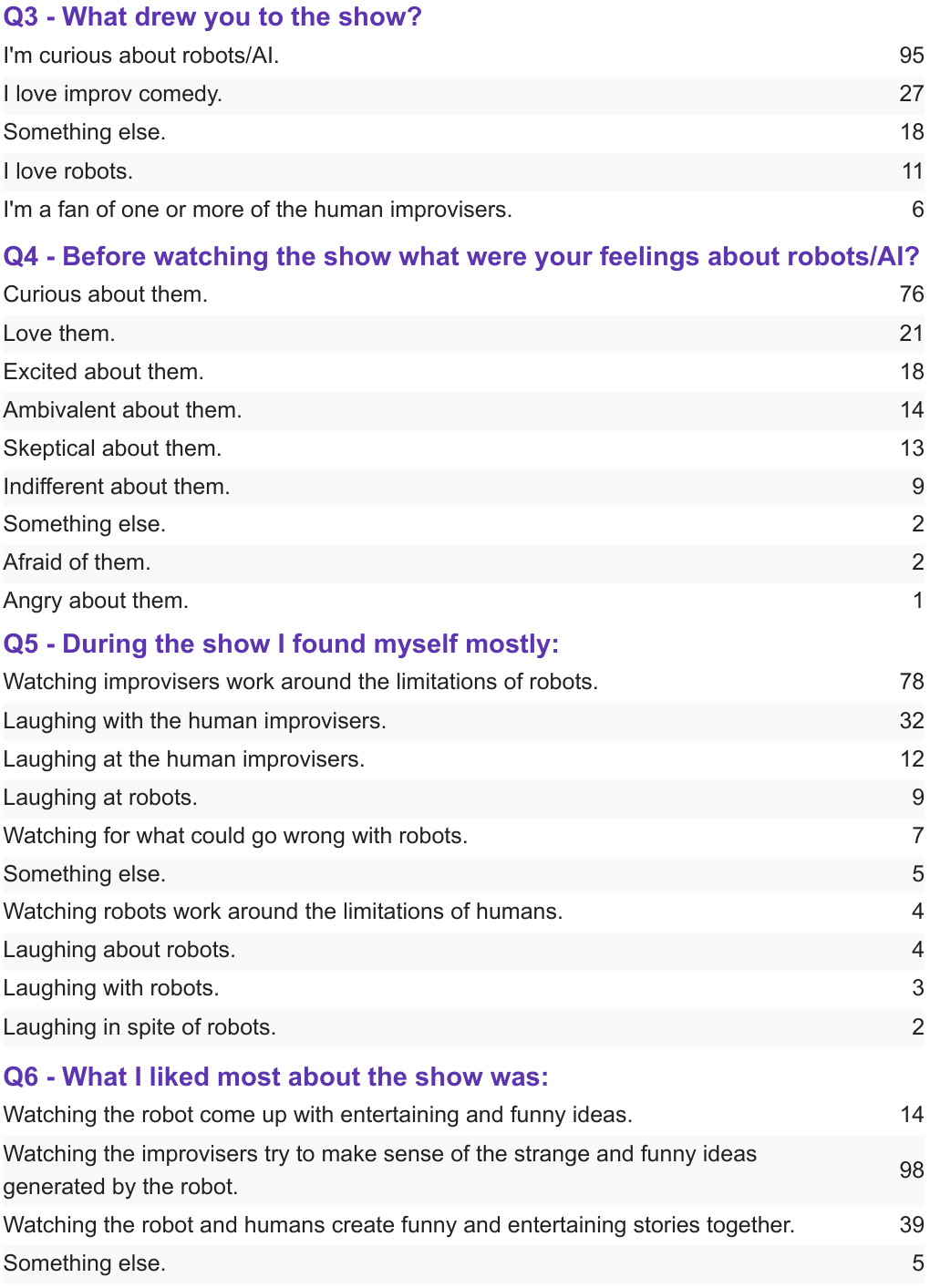}
 \includegraphics[width=0.49\columnwidth]{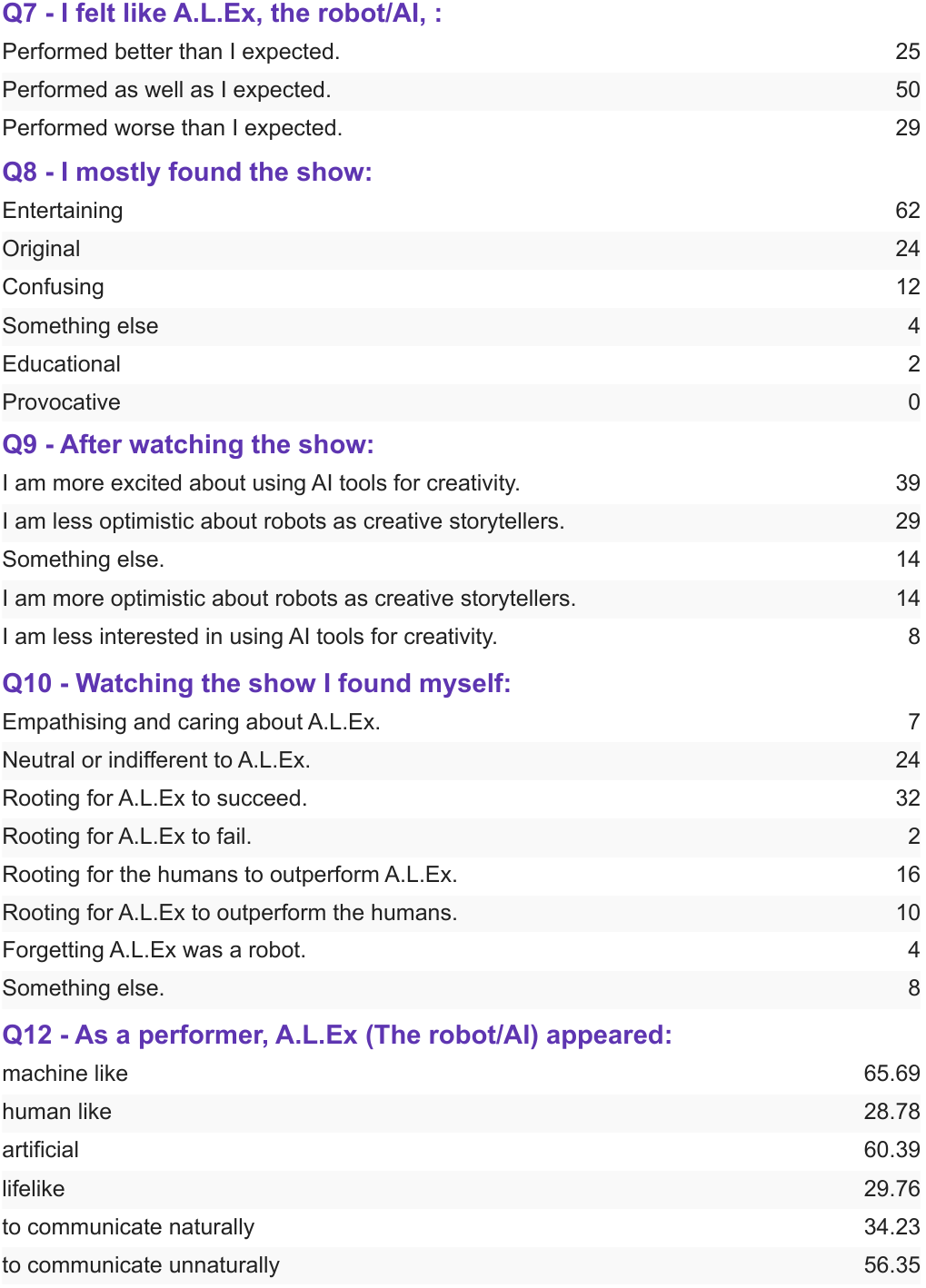}
 \caption{Audience survey results.}
 \label{fig:audience-results}
\end{figure}

\begin{figure}
 \centering
 \includegraphics[align=t,width=0.49\columnwidth]{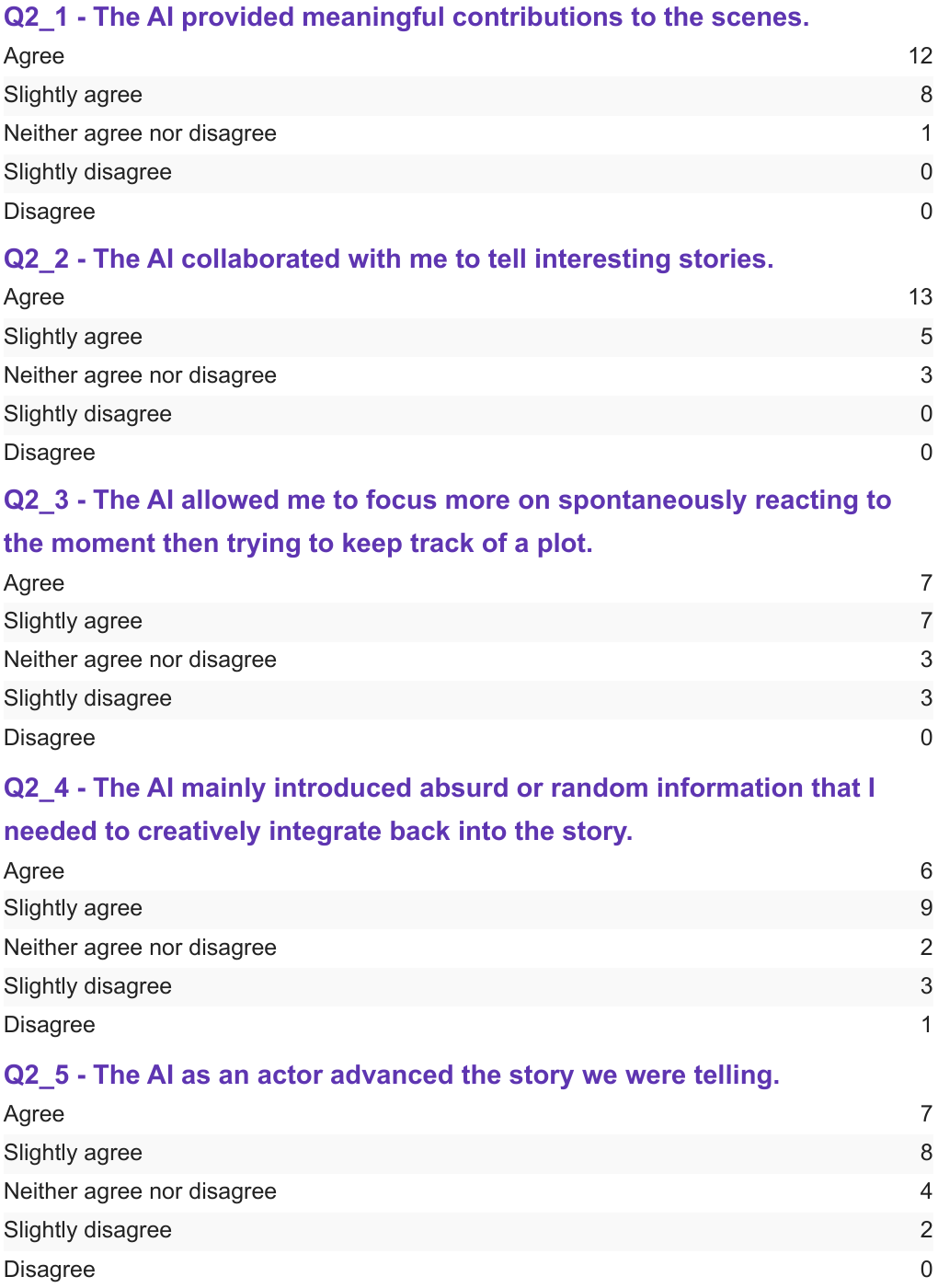}
 \includegraphics[align=t,width=0.49\columnwidth]{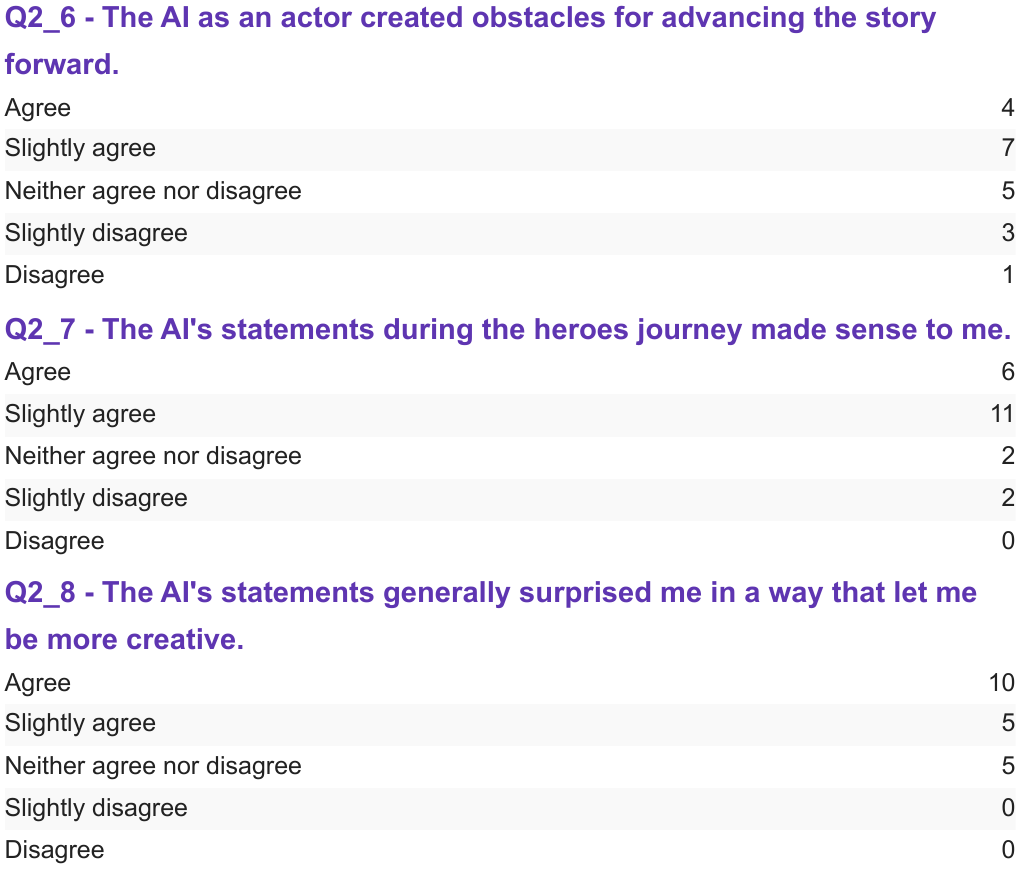}
 \caption{Actors survey results.}
 \label{fig:actor-results}
\end{figure}

Results of audience surveys from all 26 performances are shown on Figure \ref{fig:audience-results}, and of actor surveys on Figure \ref{fig:actor-results}.

\subsection{Performance of LLMs}
During the shows, the curator was presented with a stream of lines generated simultaneously in response to speech-to-text prompts by our three LLMs: Chat GPT-3.5, PaLM 2,and Llama 2 (see Fig. \ref{fig:interface-curator}). The reasons for using three LLMs were two-fold: first, this provided robustness in case one of the remote LLM services (Chat GPT-3.5 or PaLM 2) was down, and second, it provided diversity in responses due to the differences between the models (training data and model size). After analysing the dialogue system logs from all 26 performances, we noticed that LLMs were generating different numbers of lines, due to temporary model unavailability (for remotely served LLMs \emph{gpt-3.5-turbo} and \emph{text-bison}) or processing speed (for \emph{llama-2-13b-chat}, served locally). After normalising by the number of generated lines, we noticed that each LLM had, overally, a comparable chance of being selected by the curator. LLMs were not retrained on improv-specific datasets because of the late release of Llama 2 prior to the Edinburgh Festival and because fine-tuning was not availabile for Chat GPT-3.5 and PaLM 2 at that time.

\subsection{Audience Perceptions of an AI Actor}
The audience members who filled out our survey reported mixed perceptions of the AI's ``acting'' ability. According to results for Q10, the respondents were ``rooting for the AI to succeed'' more than any other option. However, results around specific acting metrics suggests that most felt that the AI's performance did not meet expectations. Q12 (109 responses) asked the audience to rate the AI's performance as a performer in several areas, including\textit{ naturalness of communication} (avg: 33), ``unique mind'' (avg: 45), ``machine-like appearance'' (avg: 64). Q14 (100 responses) asked if the AI's responses appeared to be ``similar to a human'' (avg: 53) and ``motivated toward mutual benefit with other actors'' (avg: 64). Of note is that despite the appearance of some originality, positive intent, and degree of human-like response, in the context of performance, the AI generally still presented machine-like responses perceived as ``ignorant of the scenes'' (avg: 76). We discuss later how challenges for LLMs and speech recognition to clearly differentiate between multiple speakers may contribute to this. 

\subsection{AI Partner vs AI Entertainer}
\label{sec:ai-partner}
While the AI did not appear to rate high as an improviser, more respondents reported in Q9 feeling ``excited about using AI tools for creativity'' after watching the show (37\%) than before the show (with only 16\% feeling more ``optimistic about AI as storytellers''). We discuss later how this result might inform the impact of showing AI as collaborating with humans in distinctly \textit{artificial} ways as opposed to presenting the AI in the role of a human.

\subsection{Public Perception of Conversational AI Capability}

Audiences did not perceive AI as capable of producing multi-party contextually meaningful dialogue. As summed-up by a participant: \emph{``I could see the potential of [the AI] but it was clear he still needs to improve to understand complex multi-human dialogues in rapidly changing scenarios. The voice recognition was also occasionally inputting incorrect words to [the AI] on important sentences which was limiting what he could do.''} Interestingly, while the AI's responses were far from what we expect from a human improviser, they were not perceived as completely unrelated to the context of the scenes:
more than half reported that the AI appeared \emph{``responsive to what was happening on stage''}, with nearly half reporting responses appeared \emph{``unique''} and to have their own \emph{``style''}.

By its nature, improvised theatre presents a difficult problem for conversational AI \citep{martin2016improvisational}: the roles for characters are emergent and based on shared mutual understanding of complex social structures and norms.
Theatre scenes have both tacit and explicit rules of engagement. Subsequently, attempting to build a system that can shift between any given emergent situation is still beyond the capacity of publicly available text-only chatbots.
One of the objectives of the our show 
was to contribute to the public understanding of the aforementioned problem. Looking at the actual text generated by the LLMs, despite speech recognition errors, responses demonstrated reasonable understanding of the context (see Fig: \ref{fig:dialogue}). While having a human-in-the-loop to provide contextual data for the prompting results in better possible responses, the human curators of the responses did not consistently select the most appropriate ones: this is understandable as the human curator was tasked with both listening to what is being performed on stage as well as reading from a constant stream of available lines, appearing with a delay. Furthermore, the Cyborg hearing the line would often need to delay to speak the line as they waited for an appropriate opportunity in the scene. Subsequently, turn-taking remains a difficult challenge for performing with AI.      

\subsection{Curiosity for Robots and AI}
The results of Q3 shows that ``curiosity for robots/AI'' was the leading driver of audiences, significantly overtaking ``love for improv comedy'' in general, or ``love for robots'' in general, or simply ``being a fan of the show or a cast member''. This result supports the idea that we are in a transitional phase of accepting AI into mainstream cultural experience.
This reason for going to the show correlates with answers to Q4, namely the vast majority of audiences being more ``curious about robots'' than any other feeling including ``skepticism'', ``excitement'', ``fear'', ``indifference'', or ``love'' for them. The show had positive audience reviews\footnote{e.g, \emph{``thoroughly enjoyable, funny but at the same time there were some thought provoking moments'', ``I think the actors did a great job at working with the AI. I especially liked the cyborg concept of an AI driving the word choices of a character'', ``Original improv show which educates and involves the audience. Entertaining and great for families with older kids'', ``Great concept for a show to provide the improv actors with good source material… and these actors were very good at developing chosen themes to conclusion'', ``Anything can happen. Very unusual but it really works''}
\scriptsize \url{tickets.edfringe.com/whats-on/artificial-intelligence-improvisation}
} and mixed press reviews, with several prominent publications outright criticising it for not being entertaining like a typical comedy show\footnote{\emph{``[...] the most interesting parts of the show come when the AI (represented on stage by either a tiny robot or a cast member wearing an earpiece) is allowed to strut its stuff. Though most of what it comes out with is an odd combination of intensely logical and amusingly incoherent, it’s still fascinating to watch a computer, on a basic level, make jokes and quips in response to a number of quirky scenarios. Unfortunately, those moments make up nowhere near enough of the show, which can’t seem to decide whether its purpose is to inform or to entertain. [...] By the end, it’s troubling to admit that almost all of the laughs in the show came not from the human cast but from the AI robot facing them. And though a final live recreation of the Turing test provides an intriguing end to the show, ironically, a more in-depth explanation of the science behind AI would have been both more entertaining and more interesting.''}
\scriptsize \url{whynow.co.uk/read/} \url{artificial-intelligence-improvisation-review-comedy} \url{-has-nothing-to-fear-from-ai}
}, while others outright praising it for its high level of entertainment and originality\footnote{\emph{``A fantastic look at how AI can be used in an artistic space while also showing us that it isn’t quite as advanced as we all feared, as it struggles with even the most mundane task, like generating dialogue for an investment banker who wishes to become an astronaut, you know, everyday things like that. It may be able to fool university examiners, but sadly it’s not quite ready to fully embody a midlife crisis.
Thoroughly enjoyed the show, it was interesting to see just how well the AI prompts were able to keep up with each of the ongoing scenes, or not as was sometimes the case, and how the rest of the cast had to react to this. Certainly different, definitely entertaining.''}
\scriptsize \url{https://theatreandtonic.co.uk/blog/artificial-intelligence-improvisation-review}
}. The self-perception by the creative team of the show was that we needed to juggle advocacy for the use of AI as a tool for creativity, public communication about risks and potentials of AI, high audience expectations about the capabilities of AI, and the need to make a comedic and entertaining show. Of note is that the show was a commercial success in an environment where commercial success was not guaranteed: the audience attendance stayed consistent throughout the run, averaging 63\% seat capacity in a 106-seat venue. This curiosity is likely a result of the saturation of news about advances in AI, and their increased use by the general public.

\subsection{Multi-Party Dialogue With Human-Curated AI}
As discussed in the previous section,
the majority of audiences reported feeling ``more excited about using AI as a creative partner'', while reporting being ``less optimistic about AI as a creative storyteller''. We interpret this as a promising avenue for (human-centered) human-machine collaboration.

Our use of human-in-the-loop curation and prompt engineering likely contributed to this result. By using a human curator, we work outside of the traditional thinking around programming AI for social interaction: we take a system designed for two-party dialogue and try to make it work for multi-party dialogue through use of a curator. We believe our work is more related to human augmentation with AI than the development of autonomous AI.

\subsection{Making Limited Robots Subjects of Performance}

While audiences reported high levels of entertainment and satisfaction about the overall experience, they did not experience an immediate enhancement of improvisers: in the survey, they reported that the robot sometimes ``gets in the way'' of good improvisation: ``During the show, I found myself mostly
watching improvisers work around the limitations of robots'' (Q5), as opposed to  ``laughing at'', ``with'', or ``about'' robots.
On one hand, this seems to indicate that the staged robot was seen as an obstacle more than an equal partner on stage, and that our staging of LLMs in multi-party dialogue did not consistently produce convincing human-like multi-party chat. On the other hand, this suggests that the majority responded in a way that kept the robot as subject rather than object.
This is further supported with statements from the audience about what they liked best about the show: \emph{``how interaction between robot and human works''}, \emph{``I wasn't entirely sure which bits were the robots and which bits were the humans. It wasn't always clear!''}, \emph{``Abstract thinking and creations of robots and creative interpretation of robot thoughts (cyborg) and interactions between humans and so''} and \emph{``Watching the improvisers try to make sense of the strange and funny ideas generated by the AI.''}
40 participants (out of 150) reported they enjoyed \emph{``watching the robot and humans create funny and entertaining stories together.''} We extrapolate from the responses a keen interest from the audience for seeing how to engage a robot, as much if not more than seeing to what degree an artificial intelligence can pass for human intelligence. As discussions around the role of AI in the entertainment industry are often concentrated around human replacement, our audience observations highlight how AI can also be appreciated in entertainment as ontologicially independent objects in dialogue with humans. We explore this subject further in a discussion of applications of our findings for future research.

\section{Discussion and Future Work}
\subsection{Ethical Implications}

As generative AI technology developed, it became widely available to the general public.
The fact that text can be generated in the style of a specific writer gave rise to controversies and ethical concerns about plagiarism and misappropriation of artistic work that cannibalise creative economies
\citep{weidinger2021ethical,frosio2023generative,jiang2023ai}.

By employing generative AI in the context of a show for diverse theatre festival audiences, we provoked and then engaged members of the general public attending our show about their perception of generative AI. We did this by illustrating possible uses of AI, inviting their scrutiny during and after the performance, and addressing concerns of the cast members and artists with whom we discussed about the show. Specifically, we discussed the format and aim of the AI-based improvisation with our cast members, with members of the public to whom we flyered the show, with audiences in informal discussions before and after the show, with audiences during the performance through a qualitative survey, with journalists from over 10 different press venues who interviewed us (see Appendix), and with participants of a panel on art and AI during Edinburgh Festival Fringe 2023. Common concerns focused on copyright and the misappropriation of artists' work when using image generation, and its destructive impact on the creative economies. Additional concerns included the appropriateness of generated images, their multiple representational biases, and the devaluation (via automation) of human creative work.

In our show format, we presented collaborative and co-creative applications of generative AI that gave human performers agency by inviting them directly into the generation loop, curating and responding to outputs from AI as part of live interaction. As the audience could witness, results of the generative AI were not the final artistic output: they served impermanent and improvised theatre, acting as source material to inspire live performance by human actors.

\subsection{Applications for Future Research}

Staging LLMs in a noisy multi-party conversational context yielded several insights for future research around the role of AI in live entertainment. 
The LLMs we deployed appeared technically capable of generating meaningful responses in the context of multi-party dialogue \textit{when provided sufficient data}.
How best to provide sufficient data for multi-party dialogue LLMs remains an open question. 
Our approach captured live audio from performers and relied on a ``human-in-the loop'' to manually inject context and curate best responses, which introduced delays into the system. 
Multi-microphone systems supplemented by sentiment analysis on the speech tone, could reduce delays and ambiguity.


Live improvisational theatre settings provide opportunities for audiences to engage with the co-creative potential of AI beyond the logic of human replacement. Using AI to produce static content like images, videos, music or text, which are presented as de-contextualized artefacts, may raise challenges in determining authorship;
whereas in the live arts settings, the AI is allowed to appear as a (potentially anthropomorphic) subject alongside human performers, thereby showcasing collaboration. We therefore encourage more research around the perception of artificial intelligence specifically through the medium of live entertainment.

\bibliographystyle{iccc}
\bibliography{edinburgh}

\section{Acknowledgments}
We thank the cast members of Improbotics who contributed their talents and ideas in the run-up to and during the 2023 Edinburgh Festival Fringe performances, namely Alex Newson, Ben Lovell, Em Stroud, Fiona Howat, Holly Mallett, Jenny Elfving, Jillian Ellis, Jodie Irvine, Julie Flower, Marouen Mraihi, Mike Prior, Paul Little, Roel Fox, Sarah Davies, Thomas Jones, Tommy Rydling, as well as guest improvisers Calum Jarvie, Caroline Matison, Charles Dundas, Gregor Davidson, Ken Gordon, Ryan Murphy and Steven Millar. We are grateful to audiences for critical vocal feedback, and to Dr. Lidia Alvarez for her assistance with the survey analysis.

\section{Appendix}
\label{appendix}

\subsection{Technical Implementation Details}
\label{sec:technical}

We implemented the speech recognition system for the shows using a USB microphone (Blue Yeti\footnote{\scriptsize \url{https://www.logitechg.com}}) placed on a microphone stand at the front of the stage, and OpenAI's Whisper speech recognition software running continuously during the show (we used the 4-bit quantized {\tt whisper-small.en} model and a modified C++ implementation of Whisper\footnote{\scriptsize \url{https://github.com/ggerganov/whisper.cpp}} running with a typical 0.3s latency on the stage laptop, a MacBook Pro M2 with 96GB memory) that would send each line of recognised speech to the language model server.

The language model server processed each incoming line of speech-recognised text, as well as each line of context manually typed by an operator, to assemble the context prompt for the LLMs. A typical example of assembled prompt would consist of (1) a system prompt, followed by (2) lines of dialogue from speech recognition, and (3) an instruction.

For example, for ``Couples' Therapy'', the system prompt (1) was: 
\emph{You are an improv actor doing role-play with me. You stay in character and only say the lines that your character would say. You are performing for an adult audience and your goal is to entertain them with your irreverent wit. Below is the setup for an improvised scene. You work as a couple therapist and counselor. A distraught couple enters your office. You desperately try to save their relationship, but constantly give comically bad advice for humorous effect.} The prompt was then concatenated with (2) lines of dialogue coming from speech recognition, prefixed with the name of the human speaker (entered manually by the operator), and lines of dialogue generated by the LLM, e.g.:
\emph{Paul: Doctor, we need help, my partner Alex wears Birkenstocks and picks his toenails.} Finally, the prompt was concatenated with an instruction (3) such as: \emph{You play the role of Alex. Write several possible responses for Alex.}. Additional metadata of context were directly added to the dialogue section (2) of the prompt, e.g..: \emph{Alex starts speaking in a literary style and makes many funny puns.}.

The LLM server used the Google Palm 2 API and the OpenAI ChatGPT-3.5 and ChatGPT-4 API to access the remote LLMs as well as 4-bit quantized Llama 2 13B running locally on the laptop. We developed a centralised server architecture that allowed multiple computers and tablets to access the history of speech recognition results, operator-input context metadata, LLM-generated lines and lines selected by the curator. During the show, a version of the UI was projected on screen.

\subsection{Consent and Privacy Notice in Surveys}
\label{sec:consent}

Audience surveys started with:
\emph{``Thank you so much for attending the Improbotics show, 
and for taking the time to leave feedback about your experience. This survey is completely anonymous and should take about 5 minutes. Please click to 'consent' if you are happy to proceed.''} and concluded with 
\emph{``The data you provided will be used for the purposes of research by Improbotics and University of Kent. 
This survey is anonymous, please be assured that your responses will be kept completely confidential.
If you would like to contact the principal investigators in the study to discuss this research, please e-mail Dr. Boyd Branch at bmb22 [AT] kent.ac.uk, or Dr. Piotr Mirowski at piotr.mirowski [AT] computer.org.
''}

\emph{``For information about how we protect your and use your data please follow this link: \scriptsize \url{research.kent.ac.uk/ris-research-policy-support/wp-content/uploads/sites/2326/2021/06/GDPR-Privacy-Notice-Research.pdf}''}

\subsection{Audience Questionnaire}
\label{sec:audience}

\begin{itemize}
    \item Q1. (Consent and privacy notice).
    \item Q2. Which show did you attend? \\ (List of dates)
    \item Q3. What drew you to the show? (single choice) \\ I love improv comedy. / I love robots. / I'm curious about robots/AI. / I'm a fan of one or more of the human improvisers. / Something else\textsuperscript{*}.
    \item Q4. Before watching the show what were your feelings about robots/AI? (single choice) \\ Love them. / Afraid of them. / Curious about them. / Skeptical about them. / Indifferent about them. / Ambivalent about them. / Angry about them. / Excited about them. / Something else\textsuperscript{*}.
    \item Q5. During the show I found myself mostly: (single choice) \\ Laughing at robots. / Laughing at the human improvisers. / Laughing with robots. / Laughing with the human improvisers. / Laughing about robots. / Laughing in spite of robots. / Watching for what could go wrong with robots. / Watching improvisers work around the limitations of robots. / Watching robots work around the limitations of humans. / Something else\textsuperscript{*}.
    \item Q6. What I liked most about the show was: (single choice) \\ Watching the robot come up with entertaining and funny ideas. / Watching the improvisers try to make sense of the strange and funny ideas generated by the robot. / Watching the robot and humans create funny and entertaining stories together. / Something else\textsuperscript{*}.
    \item Q7. I felt like A.L.Ex, the robot/AI, : (single choice) \\ Performed better than I expected. / Performed as well as I expected. / Performed worse than I expected.
    \item Q8. I mostly found the show: \\ Entertaining / Educational / Original / Provocative / Confusing / Something else\textsuperscript{*}.
    \item Q9. After watching the show: (single choice) \\ I am more optimistic about robots as creative storytellers. / I am less optimistic about robots as creative storytellers. / I am more excited about using AI tools for creativity. / I am less interested in using AI tools for creativity. / Something else.
    \item Q10. Watching the show I found myself: (single choice) \\ Empathising and caring about A.L.Ex. / Neutral or indifferent to A.L.Ex. / Rooting for A.L.Ex to succeed. / Rooting for A.L.Ex to fail. / Rooting for the humans to outperform A.L.Ex. / Rooting for A.L.Ex to outperform the humans. / Forgetting A.L.Ex was a robot. / Something else\textsuperscript{*}.
    \item Q12. As a performer, A.L.Ex (The robot/AI) appeared: (scores between 0 and 100) \\ machine like / human like / artificial / lifelike / to communicate naturally / to communicate unnaturally / to have a mind of its own / not to have a mind of its own.
    \item Q13. A.L.Ex's responses as an independent intelligence appeared: (scores between 0 and 100) \\ fake/not really generated in response to what was happening on stage. / real/ actually responsive to what was happening on stage. / common/generic / unique / a sham/ not really AI / run of the mill/ ordinary / original/ distinct / to have their own style.
    \item Q14. A.L.Ex's responses as an actor/improviser appeared: (socres between 0 and 10) \\ similar to a human actor's responses. \ reciprocal/ motivated toward mutual benefit with other actors. \ supportive of the scenes \ ignorant of the scenes.
    \item Q15-Q19. Please elaborate (in case ``Something else'' was chosen on questions Q3, Q4, Q5, Q8 and Q10).
    \item Q20. Is there anything you would like to share with us about your experience watching the show?   
\end{itemize}

\subsection{Actor Questionnaire}
\label{sec:actor}

\begin{itemize}
    \item Q1. (Consent and privacy notice).
    \item Q2. While improvising with A.L.Ex \\ A.L.Ex provided meaningful contribtions to the scenes. / A.L.Ex collaborated with me to tell interesting stories. / A.L.Ex allowed me to focus more on sponteously reacting to the moment then trying to keep track of a plot. / A.L.Ex mainly introduced absurd or random information that I needed to creatively integrate back into the story. / A.L.EX as an actor advanced the story we were telling. / A.L.Ex as an actor created obstacles for advancing the story forward. / A.L.Ex's statements during the heroes journey made sense to me. / A.L.Ex's statements generally surprised me in a way that let me be more creative. / A.L.Ex's statements surprised me in a way that made it more difficult for me to understand the story I was a part of. / I enjoyed performing with A.L.Ex
    \item Q3. Please describe what it was like to perform a long form scene with A.L.Ex.
    \item Q7. Please describe what it was like to perform short form games with A.L.Ex.
    \item Q4. Please describe how the choices A.L.Ex made as an improviser impacted your performance as an improviser.
    \item Q5. Please describe the biggest challenges for you with A.L.Ex as a scene partner.
    \item Q6. Please describe what you enjoyed most about having A.L.Ex as a scene partner.
\end{itemize}

\subsection{Press Interviews}
\label{sec:press}

Improbotics was interviewed by Tina Daheley for the {\it BBC World Service - Cultural Frontline} in ``What the AI revolution means for arts'', published 4 March 2023\footnote{\scriptsize \url{https://www.bbc.co.uk/programmes/w3ct37sv}},
by Mike O'Sullivan for {\it Voice of America} in ``Artificial Intelligence Can Create, But Lacks Creativity, Say Critics'', publishhed on 26 April 2023\footnote{\scriptsize \url{https://www.voanews.com/a/artificial-intelligence-can-create-but-lacks-creativity-} \url{say-critics/7068177.html}}, 
by Gary Baum for the {\it Hollywood Reporter} in ``Why AI Isn’t Funny: At Least Not Yet'', published on 1 June 2023\footnote{\scriptsize \url{https://www.hollywoodreporter.com/business/digital/why-ai-isnt-funny-at-least-not-yet-1235503678/}}, by Jay Richardson for {\it The Scotsman} in ``AI is taking over Fringe comedy; can robots be funnier than humans?'', published on 31 July 2023\footnote{\scriptsize \url{https://www.scotsman.com/arts-and-culture/edinburgh-festivals/ai-is-taking-over-fringe-comedy-can} \url{-robots-be-funnier-than-humans-4238383}}, 
by Elizabeth Greenberg for {\it DIGIT News} in ``Yes-anding AI: Artificial Intelligence Stars at the Edinburgh Fringe'', published on 16 August 2023\footnote{\scriptsize \url{https://www.digit.fyi/yes-anding-ai-artificial} \url{-intelligence-stars-at-the-edinburgh-fringe/}}, 
by Gillian Tett for the {\it Financial Times} in ``Can AI crack comedy?'', pubkished on 26 August 2023\footnote{\scriptsize \url{https://www.ft.com/content/818f2cab-57ff-42c3-917b-4a83f1d87802}}, by Katie Collins for {\it CNET} in ``AI Took the Stage at the World's Largest Arts Festival. Here's What Happened'', published on 2 September 2023\footnote{\scriptsize \url{https://www.cnet.com/tech/ai-took-the-stage-at-the-worlds-largest-arts-festival} \url{-heres-what-happened/}}.

\subsection {Actor Full Responses}
\label{sec:survey}

\subsubsection{Please describe what it was like to perform a long form scene with A.L.Ex.}

\emph{
\begin{itemize}
    \item ALEx was able to respond in context, with puns. At one point it was even able to rap far better than I could under pressure
    \item The Heroes Journey I performed with Alex was notable for one scene where the lines were completely on point and kept the scene moving along. But these lines were embodied brilliantly by the performer who was playing the cyborg. The lines were good by A.L.Ex but made excellent by the performance.
    \item Creative and exciting
    \item Complex
    \item I was the cyborg being fed lines by A.L.Ex in the long form. I felt the lines were very relevant to what was being said. However, I felt a vital piece of info was just out of reach of A.L.ex and the human improvisers that would have given the long form its ending / resolve that it felt I needed. Not being able to give it as an improviser I felt frustrated and helpless to assist those I was performing with.
    \item Mad and silly
    \item Hero’s journey was real fun today
    \item This show was a lot more interesting today as we had the improviser using a lot more physicality.
    \item It was interesting to perform knowing that there is a human choosing the line for A.L.Ex to say. And that allowed them choose lines that keep the plot on slightly together whilst allowing for moments of absurdity
    \item In the long form scene Hero's Journey A.L.Ex did produce more lines that were of a absurd or non-sequitur variety that as performers we had to integrate into the story. This wasn't a bad thing - the audience very much enjoyed seeing us performer's struggle and adapt to that - but it did mean we had to steer the plot more than in previous shows.
    \item Surprising, varied, harder work than with a human to progress the story or add richness to the relationship
    \item It adds a different and surprising element to the scenes.
    \item Can be tough to move the scene on
    \item It was a bit of a slog. A.L.Ex did produce some nice moments with the performers but felt majority of the time was not giving performers much to work with in regards to funny lines or lines to progress the story.
\end{itemize}
}

\subsubsection{Please describe how the choices A.L.Ex made as an improviser impacted your performance as an improviser.}

\emph{
\begin{itemize}
    \item ALEx was like another performer, often responses elevated the scene. Less for when lines generated were fairly generic or it the language model did not return a response
    \item As A.L.Ex is unchangeable in its performance (could that be seen as a choice?) I have to be aware of my choices in my performance. Not sure that makes sense.
    \item Pushed me to take more risks
    \item He makes it difficult to yes and
    \item I had to adapt the  character I was playing in the dating scene. I came on as a high status character but then had my status lowered by a comment by A.L.ex. It reminded me in the importance to be able to adapt status as an imoroviser. Be aware of it.
    \item Like doing long form with a 5 year old
    \item It helped shape the story or take it into directions I wasn’t expecting! But that is also due to the human selecting the lines
    \item I think sometimes people would have to justify what A.L.Ex had said in the scene depending on how the human delivered the lines. Which can sometimes create great moments or slight awkwardness
    \item I had to have more of a mind on plat and keeping the scenes grounded in the realiaty we had created.
    \item I had to do more on plot, relationship and justifying
    \item Having puns for too long meant i had to change modes, for the ted talk it enabled true fun and in the moment thinking.
    \item Helps me to listen more
    \item It made it harder to be in the moment this show. This maybe because we were a small cast so when we weren't improvising we had to select the lines from the ipad. This meant we were never able to watch from the sidelines to think about the scene.
\end{itemize}
}

\subsubsection{Please describe the biggest challenges for you with A.L.Ex as a scene partner.}

\emph{
\begin{itemize}
    \item Too many lines generated, not allowing scene to build. Careful selection needed by the person holding the tablet.
    \item Adapting my performance when A.L.Ex is just the robot.
    \item Just incorporating his responses
    \item Not regretting Alex’s lines
    \item Adapting to the sometimes snarky "character or tone" taken by A.L.ex.
    \item When the tech fails and Alex takes longer to respond
    \item The waiting for lines as it generates new lines for the scene
    \item If it’s early on in a scene it is that initial moment of feeling like having to fill air time to get A.L.Ex on track and into the scene
    \item There was much more justifying of A.L.Ex's lines in this show due to the increased non-sequiters or odd replies, statements
    \item Lack of complete collaboration and ‘yes, and’
    \item The time delay is still a challenge as i dont think i have slowed myself down enough to make it seemless.
    \item Keeping up with Alex
\end{itemize}
}

\subsubsection{Please describe what you enjoyed most about having A.L.Ex as a scene partner.}

\emph{
\begin{itemize}
    \item The word play and puns. The fact it could out rap me. Plus even the unrelated, most out of context lines, could be used in a humorous way
    \item The flow of the scene is increasingly easier maintain.
    \item The surprises!
    \item Mixing it up baby!
    \item The creation of a moment unsuspected.
    \item The change in energy in the scene
    \item The assured humour it can bring that get the audience
    \item The moments of the absurd lines that worked within context but provided great humour
    \item The justifying of the lines meant as performers we had to be increasingly on the same page with each other to justify what A.L.Ex said in context of the scene / show.
    \item Watching the actor playing ALEx try to incorporate the lines into a meaningful and congruent charactee
    \item Adds whole other dimension to plsy
    \item The change in energy and suggestions
    \item There were some moments that A.L.Ex produced that were nice.
\end{itemize}
}

\subsubsection{Please describe what it was like to perform short form games with A.L.Ex}

\emph{
\begin{itemize}
    \item Sharp punchy lines were available most of the time. Not always.
    \item I'm beginning to feel a marked difference in performing with A.L.Ex as a robot and performing with A.L.Ex embodied as a cyborg in these scenes. The "dating" A.L.Ex requires different performance elements. Can't quite specify what these are at the moment.
    \item Hilarious and challenging
    \item Enjoyable and silly
    \item In one scene (couples therapy) it really felt as an improviser I, A.L.ex via the human cyborg, and the audience were all invested and on the same page. And there was an anticipation about what was to come next.
    \item Like talking to a 5 year old
    \item Fun! The speed dating game works quite well with Alex
    \item It was fun as it allowed for a fresh new way to do the games, like in speed dating it makes for great contrast against strong characters.
    \item I'm taking that perfomring with A.L.Ex includes controlling / curating the lines. I was curating the lines in the short form dating scenes. I found I was stepping on the other performers lines by pressing the line too quickly so A.L.Ex cut them off. This was a result I believe from being partly used to feeding lines to the human cyborg where this teqnique is fiine because only the person as the cyborg hears it. But also partly anxiety about leaving a too big a gap between peformer speaking and A.L.Ex replying.
    \item Fun and lots of laughs
    \item Fun, challenging and random at points
    \item Exciting
\end{itemize}
}

\end{document}